\documentclass[11pt,a4paper]{article}
\usepackage[hyperref]{emnlp-ijcnlp-2019}
\usepackage{times}
\usepackage{latexsym}
\usepackage[pdftex]{graphicx}

\usepackage{url}

\aclfinalcopy % Uncomment this line for the final submission

\author{Elena \'Alvarez-Mellado\textsuperscript{1,2}, Eben Holderness\textsuperscript{1,2}, Nicholas Miller\textsuperscript{1,2}, Fyonn Dhang\textsuperscript{1}, Philip Cawkwell\textsuperscript{1},  \\
{\bf Kirsten Bolton\textsuperscript{1}},  {\bf James Pustejovsky\textsuperscript{2}}, \and {\bf Mei-Hua Hall\textsuperscript{1}} \\
\textsuperscript{1}Psychosis Neurobiology Laboratory, McLean Hospital, Harvard Medical School \\ \textsuperscript{2}Department of Computer Science, Brandeis University \\ {\tt \{ealvarezmellado, eholderness, mhall\}@mclean.harvard.edu}  \\ 
{\tt nicholas.anthony.miller@gmail.com}\\ {\tt \{pcawkwell, fdhang, kbolton\}@partners.org} \\ {\tt \{jamesp\}@cs.brandeis.edu}\\}

\title{Assessing the Efficacy of Clinical Sentiment Analysis and Topic Extraction in Psychiatric Readmission Risk Prediction}

\date{9/25/2019}

\begin{document}
\maketitle
\begin{abstract}
Predicting which patients are more likely to be readmitted to a hospital within 30 days after discharge is a valuable piece of information in clinical decision-making. Building a successful readmission risk classifier based on the content of Electronic Health Records (EHRs) has proved, however, to be a challenging task. Previously explored features include mainly structured information, such as sociodemographic data, comorbidity codes and physiological variables. In this paper we assess incorporating additional clinically interpretable NLP-based features such as topic extraction and clinical sentiment analysis to predict early readmission risk in psychiatry patients. 

\end{abstract}

\section{Introduction and Related Work}

Psychotic disorders affect approximately 2.5-4\% of the population \citep{perala2007lifetime} \citep{bogren2009common}. They are one of the leading causes of disability worldwide \citep{vos2015global} and are a frequent cause of inpatient readmission after discharge \citep{wiersma1998natural}. Readmissions are disruptive for patients and families, and are a key driver of rising healthcare costs \citep{mangalore2007cost} \citep{wu2005economic}. Assessing readmission risk is therefore critically needed, as it can help inform the selection of treatment interventions and implement preventive measures.

Predicting hospital readmission risk is, however, a complex endeavour across all medical fields. Prior work in readmission risk prediction has used structured data (such as medical comorbidity, prior hospitalizations, sociodemographic factors, functional status, physiological variables, etc) extracted from patients' charts \citep{kansagara2011risk}. NLP-based prediction models that extract unstructured data from EHR have also
been developed with some success in other medical fields \citep{murff2011automated}. In Psychiatry, due to the unique characteristics of medical record content (highly varied and context-sensitive vocabulary, abundance of multiword expressions, etc), NLP-based approaches have seldom been applied \citep{vigod2015readmit, tulloch2016exploring, greenwald2017novel} and strategies to study readmission risk factors primarily rely on clinical observation and manual review \citep{olfson1999assessing} \citep{lorine2015risk}, which is effort-intensive, and does not scale well.

In this paper we aim to assess the suitability of using NLP-based features like clinical sentiment analysis and topic extraction to predict 30-day readmission risk in psychiatry  patients.   We begin by describing the EHR corpus that was created using in-house data to train and evaluate our models. We then present the NLP pipeline for feature extraction that was used to parse the EHRs in our corpus. Finally, we compare the performances of our model when using only structured clinical variables and when incorporating features derived from free-text narratives.

\section{Data} 

The corpus consists of a collection of 2,346 clinical notes (admission notes, progress notes, and discharge summaries), which amounts to 2,372,323 tokens in total (an average of 1,011 tokens per note). All the notes were written in English and extracted from the EHRs of 183 psychosis patients from McLean Psychiatric Hospital in Belmont, MA, all of whom had in their history at least one instance of 30-day readmission. 

The age of the patients ranged from 20 to 67 (mean = 26.65, standard deviation = 8.73). 51\% of the patients were male. The number of admissions per patient ranged from 2 to 21 (mean = 4, standard deviation = 2.85). Each admission contained on average 4.25 notes and 4,298 tokens. In total, the corpus contains 552 admissions, and 280 of those (50\%) resulted in early readmissions. 

\section{Feature Extraction}
The readmission risk prediction task was performed at the admission level. An admission consists of a collection of all the clinical notes for a given patient written by medical personnel between inpatient admission and discharge. Every admission was labeled as either `readmitted' (i.e. the patient was readmitted within the next 30 days of discharge) or `not readmitted'. Therefore, the classification task consists of creating a single feature representation of all the clinical notes belonging to one admission, plus the past medical history and demographic information of the patient, and establishing whether that admission will be followed by a 30-day readmission or not. 

45 clinically interpretable features per admission were extracted as inputs to the readmission risk classifier. These features can be grouped into three categories (See Table \ref{tab:features} for complete list of features): 
\begin{itemize}
    \item [-]Sociodemographics: gender, age, marital status, etc.
    \item [-]Past medical history: number of previous admissions, history of suicidality, average length of stay (up until that admission), etc.
    \item [-]Information from the current admission: length of stay (LOS), suicidal risk, number and length of notes, time of discharge, evaluation scores, etc. 
\end{itemize}

\begin{figure*}
\centering
\includegraphics[width=0.97\textwidth]{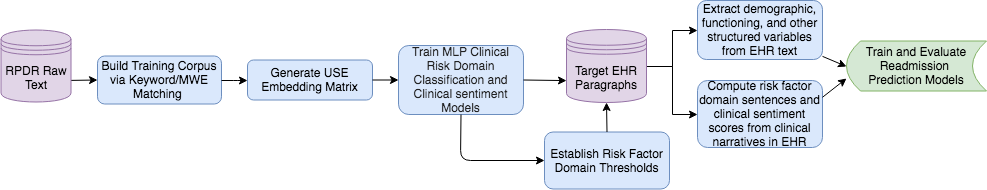}
\caption{NLP pipeline for feature extraction.}
\label{pipeline}
\end{figure*}

\begin{table}[t!]
\small
\begin{center}
\begin{tabular}{|l|}
\hline \bf Sociodemographics \\ \hline
Age\\
Gender\\
Race\\
Marital status\\
Veteran\\
\hline \bf Past medical history \\ \hline
History of Suicidality \\
Number of past admissions\\
Average length of stay (previous) \\
Average \# days between admissions \\
Previous 30-day readmission (Y/N)\\
Number of past readmissions \\
Readmission ratio \\
Average GAF at admission \\
Average GAF at discharge \\
Mode of past insight values \\
Mode of past medication compliance \\
\hline \bf Current admission \\ \hline
\bf Structured features \\
Number of notes \\
Number of tokens \\
Number of tokens in discharge summary \\
Average note length \\
GAF at admission \\
GAF at discharge \\
GAF admission/discharge difference \\
Mean GAF (all notes for visit) \\
Insight (good, fair, poor) \\
Medication Compliance \\
Estimated length of stay \\
Actual length of stay \\
Difference b/w Estimated \& Actual LOS \\
Is first admission (Y/N) \\ \hline
\bf Unstructured features \\
Number of sentences (Appearance) \\
Number of sentences (Mood) \\
Number of sentences (Thought Content) \\
Number of sentences (Thought Process) \\
Number of sentences (Substance Use) \\
Number of sentences (Interpersonal) \\
Number of sentences (Occupation) \\
Clinical sentiment (Appearance) \\
Clinical sentiment (Mood) \\
Clinical sentiment (Thought Content) \\
Clinical sentiment (Thought Process) \\
Clinical sentiment (Substance Use) \\
Clinical sentiment (Interpersonal) \\
Clinical sentiment (Occupation) \\

\hline
\end{tabular}
\end{center}
\caption{\label{tab:features}Extracted features by category.}
\end{table}

The Current Admission feature group has the most number of features, with 29 features included in this group alone. These features can be further stratified into two groups: `structured' clinical features and  `unstructured' clinical features.

\subsection{Structured Features}
Structure features are features that were identified on the EHR using regular expression matching and include rating scores that have been reported in the psychiatric literature as correlated with increased readmission risk, such as \textit{Global Assessment of Functioning}, \textit{Insight} and \textit{Compliance}:

{\bf Global Assessment of Functioning (GAF)}: The psychosocial functioning of the patient ranging from 100 (extremely high functioning) to 1 (severely impaired) \cite{aas2011guidelines}.

{\bf Insight}: The degree to which the patient recognizes and accepts his/her illness (either \textit{Good}, \textit{Fair} or \textit{Poor}).

{\bf Compliance}: The ability of the patient to comply with medication and to follow medical advice (either \textit{Yes}, \textit{Partial}, or \textit{None}).

These features are widely-used in clinical practice and  evaluate the general state and prognosis of the patient during the patient's evaluation.

\subsection{Unstructured Features}
Unstructured features aim to capture the state of the patient in relation to seven risk factor domains (Appearance, Thought Process, Thought Content, Interpersonal, Substance Use, Occupation, and Mood) from the free-text narratives on the EHR. These seven domains have been identified as associated with readmission risk in prior work \cite{holderness2018analysis}.

These unstructured features include: 1) the relative number of sentences in the admission notes that involve each risk factor domain (out of total number of sentences within the admission) and 2) clinical sentiment scores for each of these risk factor domains, i.e. sentiment scores that evaluate the patient’s psychosocial functioning level (positive, negative, or neutral) with respect to each of these risk factor domain. 

These sentiment scores were automatically obtained through the topic extraction and sentiment analysis pipeline introduced in our prior work \cite{holderness2019distinguishing} and pretrained on in-house psychiatric EHR text. In our paper we also showed that this automatic pipeline achieves reasonably strong F-scores, with an overall performance of 0.828 F1 for the topic extraction component and 0.5 F1 on the clinical sentiment component. 

The clinical sentiment scores are computed for every note in the admission. Figure \ref{pipeline} details the data analysis pipeline that is employed for the feature extraction.

%Add justification about why we also use domain sentence counts

First, a multilayer perceptron (MLP) classifier is trained on EHR sentences (8,000,000 sentences consisting of 340,000,000 tokens) that are extracted from the Research Patient Data Registry (RPDR), a centralized regional data repository of clinical data from all institutions in the Partners HealthCare network. These sentences are automatically identified and labeled for their respective risk factor domain(s) by using a lexicon of clinician identified domain-related keywords and multiword expressions, and thus require no manual annotation. The sentences are vectorized using the Universal Sentence Encoder (USE), a transformer attention network pretrained on a large volume of general-domain web data and optimized for greater-than-word length sequences. 

Sentences that are marked for one or more of the seven risk factor domains are then passed to a suite of seven clinical sentiment MLP classifiers (one for each risk factor domain) that are trained on a corpus of 3,500 EHR sentences (63,127 tokens) labeled by a team of three clinicians involved in this project. To prevent overfitting to this small amount of training data, the models are designed to be more generalizable through the use of two hidden layers and a dropout rate \cite{srivastava2014dropout} of 0.75.

The outputs of each clinical sentiment model are then averaged across notes to create a single value for each risk factor domain that corresponds to the patient's level of functioning on a -1 to 1 scale (see Figure 2). 

\begin{figure}
\centering
\includegraphics[width=0.45\textwidth]{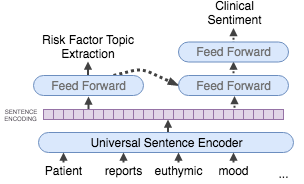}
\caption{Model architecture for USE embedding generation and unstructured feature extraction. Dotted arrows indicate operations that are performed only on sentences marked for 1+ risk factor domain(s). USE top-layer weights are fine-tuned during training.}
\label{model}
\end{figure}

\section{Experiments and Results}
We tested six different classification models: Stochastic Gradient Descent, Logistic Regression, C-Support Vector, Decision Tree, Random Forest, and MLP. All of them were implemented and fine-tuned using the scikit-learn machine learning toolkit \cite{pedregosa2011scikit}. Because an accurate readmission risk prediction model is designed to be used to inform treatment decisions, it is important in adopting a model architecture that is clinically interpretable and allows for an analysis of the specific contribution of each feature in the input. As such, we include a Random Forest classifier, which we also found to have the best performance out of the six models.

To systematically evaluate the importance of the clinical sentiment values extracted from the free text in EHRs, we first build a baseline model using the structured features, which are similar to prior studies on readmission risk prediction \citep{kansagara2011risk}. We then compare two models incorporating the unstructured features. In the "Baseline+Domain Sentences" model, we consider whether adding the counts of sentences per EHR that involve each of the seven risk factor domains as identified by our topic extraction model improved the model performance. In the "Baseline+Clinical Sentiment" model, we evaluate whether adding clinical sentiment scores for each risk factor domain improved the model performance. We also experimented with combining both sets of features and found no additional improvement. 

Each model configuration was trained and evaluated 100 times and the features with the highest importance for each iteration were recorded. To further fine-tune our models, we also perform three-fold cross-validated recursive feature elimination 30 times on each of the three configurations and report the performances of the models with the best performing feature sets. These can be found in Table \ref{tab:results}.

Our baseline results show that the model trained using only the structured features produce equivalent performances as reported by prior models for readmission risk prediction across all healthcare fields \citep{artetxe2018predictive}. The two models that were trained using unstructured features produced better results and both outperform the baseline results. The "Baseline+Clinical Sentiment" model produced the best results, resulting in an F1 of 0.72, an improvement of 14.3\% over the baseline. 

In order to establish what features were not relevant in the classification task, we performed recursive feature elimination. We identified 13 feature values as being not predictive of readmission (they were eliminated from at least two of the three feature sets without producing a drop in performance) including: all values for marital status (Single, Married, Other, and Unknown), missing values for GAF at admission, GAF score difference between admission \& discharge, GAF at discharge, Veteran status, Race, and Insight \& Mode of Past Insight values reflecting a clinically positive change (Good and Improving). Poor Insight values, however, are predictive of readmission. 

\begin{table}[t!]
\begin{center}
\begin{tabular}{|l|c|c|c|}
\hline \bf Model & \bf Acc & \bf AUC & \bf F1 \\ \hline
Baseline & 0.63 & 0.63 & 0.63  \\
Baseline+Domain Sentences & 0.69 & 0.70 & 0.69 \\
Baseline+Clinical Sentiment & 0.72 & 0.72 & 0.72  \\
\hline
\end{tabular}
\end{center}
\caption{\label{tab:results}Results (in ascending order) }
\end{table}

\section{Conclusions}
We have introduced and assessed the efficacy of adding NLP-based features like topic extraction and clinical sentiment features to traditional structured-feature based classification models for early readmission prediction in psychiatry patients. The approach we have introduced is a hybrid machine learning approach that combines deep learning techniques with linear methods to ensure clinical interpretability of the prediction model. 

Results show not only that both the number of sentences per risk domain and the clinical sentiment analysis scores outperform the structured-feature baseline and contribute significantly to better classification results, but also that the clinical sentiment features produce the highest results in all evaluation metrics (F1 = 0.72).   

These results suggest that clinical sentiment features for each of seven risk domains extracted from free-text narratives further enhance early readmission prediction. In addition, combining state-of-art MLP methods has a potential utility in generating clinical meaningful features that can be be used in downstream linear models with interpretable and transparent results. In future work, we intend to increase the size of the EHR corpus, increase the demographic spread of patients, and extract new features based on clinical expertise to increase our model performances. Additionally, we intend to continue our clinical sentiment annotation project from \cite{holderness2019distinguishing} to increase the accuracy of that portion of our NLP pipeline. 

\section{Acknowledgments}
This work was supported by a grant from the National Institute of Mental Health (grant no. 5R01MH109687 to Mei-Hua Hall). We would also like to thank the LOUHI 2019 Workshop reviewers for their constructive and helpful comments. 

\bibliography{Bibliography.bib}

\begin{thebibliography}{19}
\expandafter\ifx\csname natexlab\endcsname\relax\def\natexlab#1{#1}\fi

\bibitem[{AAS(2011)}]{aas2011guidelines}
IH~Monrad AAS. 2011.
\newblock Guidelines for rating global assessment of functioning (gaf).
\newblock \emph{Annals of general psychiatry}, 10(1):2.

\bibitem[{Artetxe et~al.(2018)Artetxe, Beristain, and
  Grana}]{artetxe2018predictive}
Arkaitz Artetxe, Andoni Beristain, and Manuel Grana. 2018.
\newblock Predictive models for hospital readmission risk: A systematic review
  of methods.
\newblock \emph{Computer methods and programs in biomedicine}, 164:49--64.

\bibitem[{Bogren et~al.(2009)Bogren, Mattisson, Isberg, and
  Nettelbladt}]{bogren2009common}
Mats Bogren, Cecilia Mattisson, Per-Erik Isberg, and Per Nettelbladt. 2009.
\newblock How common are psychotic and bipolar disorders? a 50-year follow-up
  of the lundby population.
\newblock \emph{Nordic journal of psychiatry}, 63(4):336--346.

\bibitem[{Greenwald et~al.(2017)Greenwald, Cronin, Carballo, Danaei, and
  Choy}]{greenwald2017novel}
Jeffrey~L Greenwald, Patrick~R Cronin, Victoria Carballo, Goodarz Danaei, and
  Garry Choy. 2017.
\newblock A novel model for predicting rehospitalization risk incorporating
  physical function, cognitive status, and psychosocial support using natural
  language processing.
\newblock \emph{Medical care}, 55(3):261--266.

\bibitem[{Holderness et~al.(2019)Holderness, Cawkwell, Bolton, Pustejovsky, and
  Hall}]{holderness2019distinguishing}
Eben Holderness, Philip Cawkwell, Kirsten Bolton, James Pustejovsky, and
  Mei-Hua Hall. 2019.
\newblock Distinguishing clinical sentiment: The importance of domain
  adaptation in psychiatric patient health records.
\newblock In \emph{Proceedings of the 2nd Clinical Natural Language Processing
  Workshop}, pages 117--123.

\bibitem[{Holderness et~al.(2018)Holderness, Miller, Bolton, Cawkwell, Meteer,
  Pustejovsky, and Hua-Hall}]{holderness2018analysis}
Eben Holderness, Nicholas Miller, Kirsten Bolton, Philip Cawkwell, Marie
  Meteer, James Pustejovsky, and Mei Hua-Hall. 2018.
\newblock Analysis of risk factor domains in psychosis patient health records.
\newblock In \emph{Proceedings of the Ninth International Workshop on Health
  Text Mining and Information Analysis}, pages 129--138.

\bibitem[{Kansagara et~al.(2011)Kansagara, Englander, Salanitro, Kagen,
  Theobald, Freeman, and Kripalani}]{kansagara2011risk}
Devan Kansagara, Honora Englander, Amanda Salanitro, David Kagen, Cecelia
  Theobald, Michele Freeman, and Sunil Kripalani. 2011.
\newblock Risk prediction models for hospital readmission: a systematic review.
\newblock \emph{Jama}, 306(15):1688--1698.

\bibitem[{Lorine et~al.(2015)Lorine, Goenjian, Kim, Steinberg, Schmidt, and
  Goenjian}]{lorine2015risk}
Kim Lorine, Haig Goenjian, Soeun Kim, Alan~M Steinberg, Kendall Schmidt, and
  Armen~K Goenjian. 2015.
\newblock Risk factors associated with psychiatric readmission.
\newblock \emph{The Journal of nervous and mental disease}, 203(6):425--430.

\bibitem[{Mangalore and Knapp(2007)}]{mangalore2007cost}
Roshni Mangalore and Martin Knapp. 2007.
\newblock {Cost of schizophrenia in England.}
\newblock \emph{The journal of mental health policy and economics},
  10(1):23--41.

\bibitem[{Murff et~al.(2011)Murff, FitzHenry, Matheny, Gentry, Kotter, Crimin,
  Dittus, Rosen, Elkin, Brown et~al.}]{murff2011automated}
Harvey~J Murff, Fern FitzHenry, Michael~E Matheny, Nancy Gentry, Kristen~L
  Kotter, Kimberly Crimin, Robert~S Dittus, Amy~K Rosen, Peter~L Elkin,
  Steven~H Brown, et~al. 2011.
\newblock Automated identification of postoperative complications within an
  electronic medical record using natural language processing.
\newblock \emph{Jama}, 306(8):848--855.

\bibitem[{Olfson et~al.(1999)Olfson, Mechanic, Boyer, Hansell, Walkup, and
  Weiden}]{olfson1999assessing}
Mark Olfson, David Mechanic, Carol~A Boyer, Stephen Hansell, James Walkup, and
  Peter~J Weiden. 1999.
\newblock Assessing clinical predictions of early rehospitalization in
  schizophrenia.
\newblock \emph{The Journal of nervous and mental disease}, 187(12):721--729.

\bibitem[{Pedregosa et~al.(2011)Pedregosa, Varoquaux, Gramfort, Michel,
  Thirion, Grisel, Blondel, Prettenhofer, Weiss, Dubourg
  et~al.}]{pedregosa2011scikit}
Fabian Pedregosa, Ga{\"e}l Varoquaux, Alexandre Gramfort, Vincent Michel,
  Bertrand Thirion, Olivier Grisel, Mathieu Blondel, Peter Prettenhofer, Ron
  Weiss, Vincent Dubourg, et~al. 2011.
\newblock Scikit-learn: Machine learning in python.
\newblock \emph{Journal of machine learning research}, 12(Oct):2825--2830.

\bibitem[{Per{\"a}l{\"a} et~al.(2007)Per{\"a}l{\"a}, Suvisaari, Saarni,
  Kuoppasalmi, Isomets{\"a}, Pirkola, Partonen, Tuulio-Henriksson, Hintikka,
  Kiesepp{\"a} et~al.}]{perala2007lifetime}
Jonna Per{\"a}l{\"a}, Jaana Suvisaari, Samuli~I Saarni, Kimmo Kuoppasalmi,
  Erkki Isomets{\"a}, Sami Pirkola, Timo Partonen, Annamari Tuulio-Henriksson,
  Jukka Hintikka, Tuula Kiesepp{\"a}, et~al. 2007.
\newblock Lifetime prevalence of psychotic and bipolar i disorders in a general
  population.
\newblock \emph{Archives of general psychiatry}, 64(1):19--28.

\bibitem[{Srivastava et~al.(2014)Srivastava, Hinton, Krizhevsky, Sutskever, and
  Salakhutdinov}]{srivastava2014dropout}
Nitish Srivastava, Geoffrey Hinton, Alex Krizhevsky, Ilya Sutskever, and Ruslan
  Salakhutdinov. 2014.
\newblock Dropout: a simple way to prevent neural networks from overfitting.
\newblock \emph{The Journal of Machine Learning Research}, 15(1):1929--1958.

\bibitem[{Tulloch et~al.(2016)Tulloch, David, and
  Thornicroft}]{tulloch2016exploring}
AD~Tulloch, AS~David, and G~Thornicroft. 2016.
\newblock Exploring the predictors of early readmission to psychiatric
  hospital.
\newblock \emph{Epidemiology and psychiatric sciences}, 25(2):181--193.

\bibitem[{Vigod et~al.(2015)Vigod, Kurdyak, Seitz, Herrmann, Fung, Lin,
  Perlman, Taylor, Rochon, and Gruneir}]{vigod2015readmit}
Simone~N Vigod, Paul~A Kurdyak, Dallas Seitz, Nathan Herrmann, Kinwah Fung,
  Elizabeth Lin, Christopher Perlman, Valerie~H Taylor, Paula~A Rochon, and
  Andrea Gruneir. 2015.
\newblock Readmit: a clinical risk index to predict 30-day readmission after
  discharge from acute psychiatric units.
\newblock \emph{Journal of psychiatric research}, 61:205--213.

\bibitem[{Vos et~al.(2015)Vos, Barber, Bell, Bertozzi-Villa, Biryukov,
  Bolliger, Charlson, Davis, Degenhardt, Dicker et~al.}]{vos2015global}
Theo Vos, Ryan~M Barber, Brad Bell, Amelia Bertozzi-Villa, Stan Biryukov, Ian
  Bolliger, Fiona Charlson, Adrian Davis, Louisa Degenhardt, Daniel Dicker,
  et~al. 2015.
\newblock Global, regional, and national incidence, prevalence, and years lived
  with disability for 301 acute and chronic diseases and injuries in 188
  countries, 1990--2013: a systematic analysis for the global burden of disease
  study 2013.
\newblock \emph{The Lancet}, 386(9995):743--800.

\bibitem[{Wiersma et~al.(1998)Wiersma, Nienhuis, Slooff, and
  Giel}]{wiersma1998natural}
Durk Wiersma, Fokko~J Nienhuis, Cees~J Slooff, and Robert Giel. 1998.
\newblock {Natural course of schizophrenic disorders: a 15-year followup of a
  Dutch incidence cohort}.
\newblock \emph{Schizophrenia bulletin}, 24(1):75--85.

\bibitem[{Wu et~al.(2005)Wu, Birnbaum, Shi, Ball, Kessler, Moulis, and
  Aggarwal}]{wu2005economic}
Eric~Q Wu, Howard~G Birnbaum, Lizheng Shi, Daniel~E Ball, Ronald~C Kessler,
  Matthew Moulis, and Jyoti Aggarwal. 2005.
\newblock {The economic burden of schizophrenia in the United States in 2002}.
\newblock \emph{Journal of Clinical Psychiatry}, 66(9):1122--1129.

\end{thebibliography}
\bibliographystyle{acl_natbib}

\end{document}